\documentclass[11pt]{article}

\usepackage[final]{acl}

\usepackage{times}
\usepackage{latexsym}

\usepackage[T1]{fontenc}

\usepackage[utf8]{inputenc}

\usepackage{microtype}

\usepackage{inconsolata}

\usepackage{graphicx}

\usepackage{multirow}
    \usepackage{graphicx} 
    \usepackage{bm} 
    \usepackage{booktabs}
    \usepackage{colortbl}
\usepackage{tabularx}
\usepackage{array}
 \usepackage{amsmath}

\newcolumntype{Y}{>{\raggedright\ttfamily\arraybackslash}X}

\usepackage{xcolor}
\usepackage{makecell}
    
\usepackage{listings}
\lstdefinestyle{prompt}{
    basicstyle=\ttfamily\footnotesize,
    frame = single,
    frameround=tttt,
    backgroundcolor=\color{yellow!10},
    escapechar=\+,
    breakautoindent=false, 
    breaklines=true, 
    breakindent=0pt,
}
\lstdefinestyle{cypher}{
    basicstyle=\ttfamily\small,  
    breaklines=true,             
    frame=none                
}

\title{Achieving Precise Text-To-Cypher \\
Via Grounded Knowledge Graph Data Generation}

\author{Francesco Cazzaro\textsuperscript{1}, Jessica Lennon\textsuperscript{1}, Ariadna Quattoni\textsuperscript{1,2}\\
\textsuperscript{1}Universitat Politècnica de Catalunya, Barcelona, Spain \\
\textsuperscript{2}ICREA, Pg. Lluis Companys 23, Barcelona, Spain \\
  \texttt{francesco.cazzaro@upc.edu, jessica.theresa.lennon@estudiantat.upc.edu} \\ \texttt{aquattoni@cs.upc.edu}\\ }

\begin{document}
\maketitle
\begin{abstract}
Property Graphs are rapidly being  adopted as database frameworks for representing heterogeneous data sources. To enable precise access to the information contained in them we need conversational interfaces based on Text-To-Cypher (Text2Cypher) parsers. This paper presents an automatic synthetic data generation method that can be leveraged to fine-tune small LLMs for this task. We conduct experiments on all the major Text-To-Cypher benchmarks, demonstrating that with our synthetic data generation approach we can significantly increase the performance of small LLMs, allowing them to compete with much larger proprietary models. This means that in settings in which models must be locally deployed we can ensure data-sovereignty without sacrificing accuracy and without costly annotation campaigns.
\end{abstract}

\section{Introduction}
Property Graphs (PGs) are becoming increasingly prominent for representing complex relations between entities across domains but accessing the information contained in them requires expertise in query languages such as Cypher \citep{10.1145/3183713.3190657}. To reduce this barrier we need accurate conversational interfaces to PGs powered by reliable Text-To-Cypher parsers. One typical approach to develop such interfaces is to implement a Text-To-Cypher parser by directly prompting an LLM. However, there are scenarios in which this might not be possible. For example, this approach might not be feasible in highly regulated industries in which data sovereignty is a major priority. In such a setting, solutions based on locally deployed models are a must and small models that can provide high performance without demanding massive hardware resources are preferred. The typical approach in this context is to develop a Text-To-Cypher parser specialized for a target PG by fine-tuning an LLM using annotated data for the target graph. 

When fine-tuning models for Text-To-Cypher, acquiring annotated data is often the main bottleneck since such data is usually scarce and creating it is expensive and labor intensive. LLMs can be prompted to generate training data, but the generated data will typically lack the necessary diversity and graph coverage. To mitigate this problem synthetic data generation techniques focusing on comprehensive graph coverage have been proposed. These techniques are able to generate high coverage training data for a target PG without the need of human intervention. 

\citet{cazzaro-etal-2025-spot} proposed SPOT, a synthetic data generation approach in three stages. In the first stage it generates tree patterns from a target graph to produce grounded Cypher queries. In the second stage it uses a finite-state transducer to produce initial natural language realizations of the grounded Cypher queries (proto-NL). Finally in the last stage, the proto-NLs are paraphrased by an LLM to generate the final more fluent samples for training the Text-To-Cypher parser.

SPOT yielded encouraging results and the concept of graph driven synthetic data generation held the potential to ensure high graph coverage and overcome the training data bottleneck. However, the approach fell short on two main aspects described in their paper. The first shortcoming relates to the lack of expressivity of the Cypher generation. More precisely, there are some Cypher queries that cannot be generated by the approach, such as nested queries that make use of sub-queries, queries with complex return patterns or queries involving regular expression constraints on features, to name a few. The second shortcoming relates to problems with logical meaning preservation in the paraphrasing phase. While the paraphrasing approach worked well in general, for some patterns paraphrases failed to preserve the logical meaning of the query. When we ran preliminary experiments with SPOT across a wider range of Text-To-Cypher benchmarks we found that these limitations were indeed critical. 

This paper addresses the shortcomings of prior approaches to synthetic data generation for fine-tuning Text-To-Cypher parsers and presents CYQUARK, a method that can produce a rich set of Cypher queries, covering a wide range of the Cypher syntax. Furthermore, to handle the logical meaning preservation problem our method implements an improved generation technique and a filtering stage that can reduce the number of noisily generated samples.

We conduct a thorough experimental evaluation covering the main Text-To-Cypher benchmarks and a broad range of models. Our experiments show that small LLMs have low zero-shot performance, but by using the data automatically generated by CYQUARK it is possible to overcome this performance deficit with fine-tuning. As a result we can obtain locally deployable LLMs whose performance can be comparable to that of the most powerful proprietary closed-source systems. In brief, our main contributions are:
\begin{itemize}
    \item We develop a rich and robust synthetic data generation method and release the corresponding open-source data generation library for creating training data to fine-tune Text-To-Cypher models specialized for a target PG. We release our code and data publicly \footnote{\url{https://github.com/interact-erc/Text-To-Cypher-knowledge-graph-data-generation}}.   
\item With the data generated by CYQUARK it is possible to develop small locally deployable data-sovereign Text-To-Cypher parsers that have minimal dependence on outside sources and a lighter environmental footprint. 
\item We present extensive experimental results, covering the main Text-To-Cypher benchmarks and comparing a wide range of models. We believe that our study provides a reliable and rigorous measure of how well the Text-To-Cypher task is performed by current models.
\end{itemize}

\section{Related Work}

The growing popularity of property graphs has resulted in recent efforts to create benchmarks for Text-To-Cypher. Some works focus on adapting SQL benchmarks \citep{10.1007/978-981-99-8391-9_10, tiwari-etal-2025-auto} while others create benchmarks by converting RDF graphs into property graphs \citep{nie-etal-2022-graphq, feng-etal-2025-cypherbench}. Other works generate benchmarks synthetically using pipelines designed around LLMs \citep{zhong-etal-2025-synthet2c, chauhan-etal-2025-mind, lyu2026text2gqlbenchtextgraphquery, zhang-etal-2026-cypherturn} whereas others emphasize human involvement by having annotators create or at least verify samples \citep{10.1145/3511808.3557703, zhao2023rel2graph, cazzaro-etal-2025-zograscope, yufei-etal-2026-ngdbench, lee2026tiercache, Cavalleri2026bio2c}. \citet{ozsoy-etal-2025-text2cypher} aggregate synthetic examples from various sources. \citet{zhang-etal-2026-cyphersmith} create an instruction-tuning dataset exploiting a Text-to-SQL dataset in conjunction with LLMs.

Regarding Text-To-Cypher parsers, some approaches focus on fine-tuning \citep{10.1145/3511808.3557703, 10.1007/978-981-19-8746-5_13, tiwari-etal-2025-auto, zhong-etal-2025-synthet2c, choi2026kg2cypherdatacentricpipelinebuilding, 10.1145/3810460.3812781} while others leverage large language models \citep{li2024unioqaunifiedframeworkknowledge, 10.1145/3627673.3679713, zhou-etal-2024-r3, liang2024natnl2gqlnovelmultiagentframework, ranganath2026pipecypherautomaticenterprisebenchmark, ozsoy2026adaptivetesttimeinferencetext2cypher}. Some methods focus on frameworks for detecting and correcting failure modes \citep{chen2026cypherguarddatabasegroundedcorrectionreliable, prokopiou-etal-2026-drives, shin2026cydra, jung2026rasreflectionaugmentedscalingincontext, ozsoy2026extending}. The work most closely related to us is SPOT \citep{cazzaro-etal-2025-spot}, a data generation method for training parsers that, however, has limited Cypher expressivity and coverage.

\section{Preliminary: Sample Generation Process}

\begin{figure*}[]
  \centering
  \includegraphics[width=\linewidth]{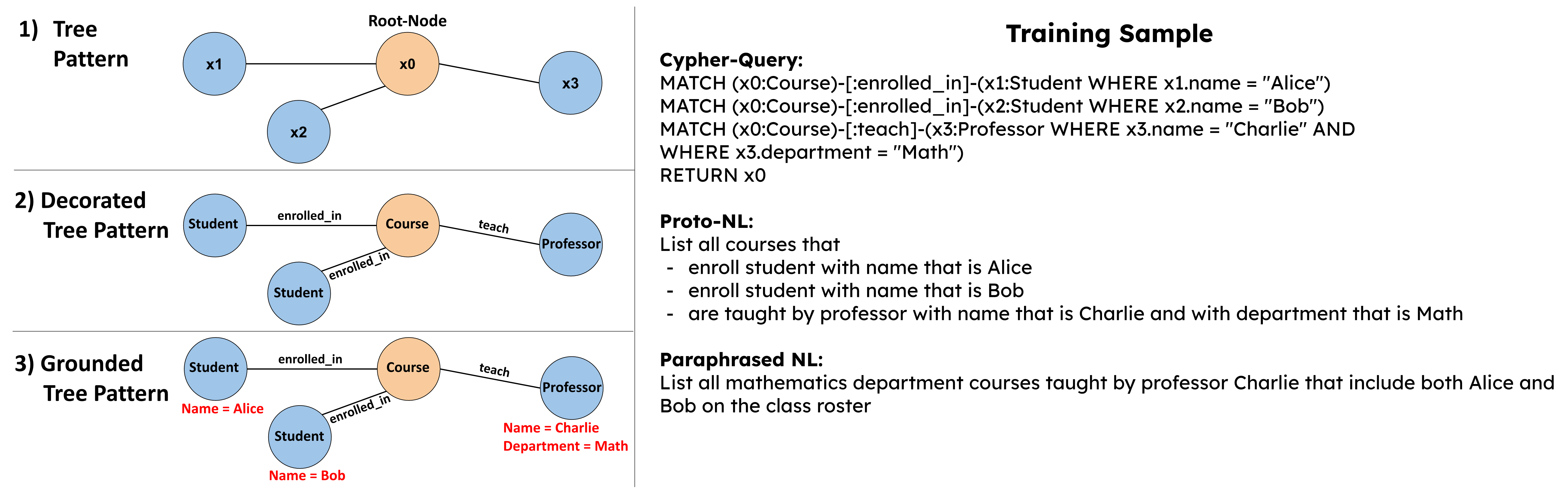}
  \caption{Query Generation Process}
  \label{fig:qp}
\end{figure*}

We start this section by providing the relevant background on property graphs and tree pattern based synthetic data generation \citep{cazzaro-etal-2025-spot}.

\subsection{Property Graphs}
A Property Graph (PG) is a graph database model where entities are represented by nodes and relationships by labeled edges. The main difference with other popular frameworks such as RDF is that in PGs both entities (i.e. nodes) and relations (i.e. edges) can have associated properties (i.e. key-value pairs). This provides a richer and more descriptive way to represent complex real-world data. PGs have gained traction in industrial applications as they enable the specification of a semantically intuitive data model in which there is a clear distinction between entity properties and entity relations. In contrast, RDF graphs conflate the two concepts since all properties must be expressed as relations. Two main languages are used for querying PGs, Cypher \citep{10.1145/3183713.3190657} and Gremlin \citep{10.1145/2815072.2815073}. Our work focuses on Cypher, the more popular among the two.
This being said, our approach could be generalized to other graph languages.

\subsection{Finite-State Transduction From Tree Patterns}

The generation process assumes two inputs: The target PG and a schema file specifying the graph’s classes, relations, and property types. For each element, the schema also provides a lexicalization field. For example, a relation named EmployeeOf connecting nodes of type Person and Company might have as lexicalization: is employed by.

The basic data creation pipeline consists of three main steps illustrated in Figure \ref{fig:qp}. The first step generates a rooted tree pattern that represents the structure of a Cypher query in which nodes and edges correspond to the entities and relations appearing in a Cypher MATCH clause. The second step decorates the tree by assigning class and relation types to each node and edge. These types are selected to be compatible with the target PG. The decorated tree pattern is then grounded by instantiating properties and other possible features with values drawn from the PG. The resulting grounded tree pattern can be deterministically mapped to a corresponding Cypher query via a finite state process. Analogously, we define a finite state transducer that generates a `proto’ natural language realization of the grounded tree pattern. The transducer that generates the proto-NL can be automatically instantiated from the graph schema. The proto-NL preserves the underlying semantics of the query but it might lack fluency and sound `artificial’; in the sense that it might not reflect a typical human language utterance. To solve this, the last step of the process uses an LLM to paraphrase the proto-NL into a fluent natural language question. This process is iterated with randomized tree patterns, decorations, and groundings, to generate a large set of question-query pairs.

\section{CYQUARK}

\begin{figure*}[h]
  \centering
  \includegraphics[width=.8\linewidth]{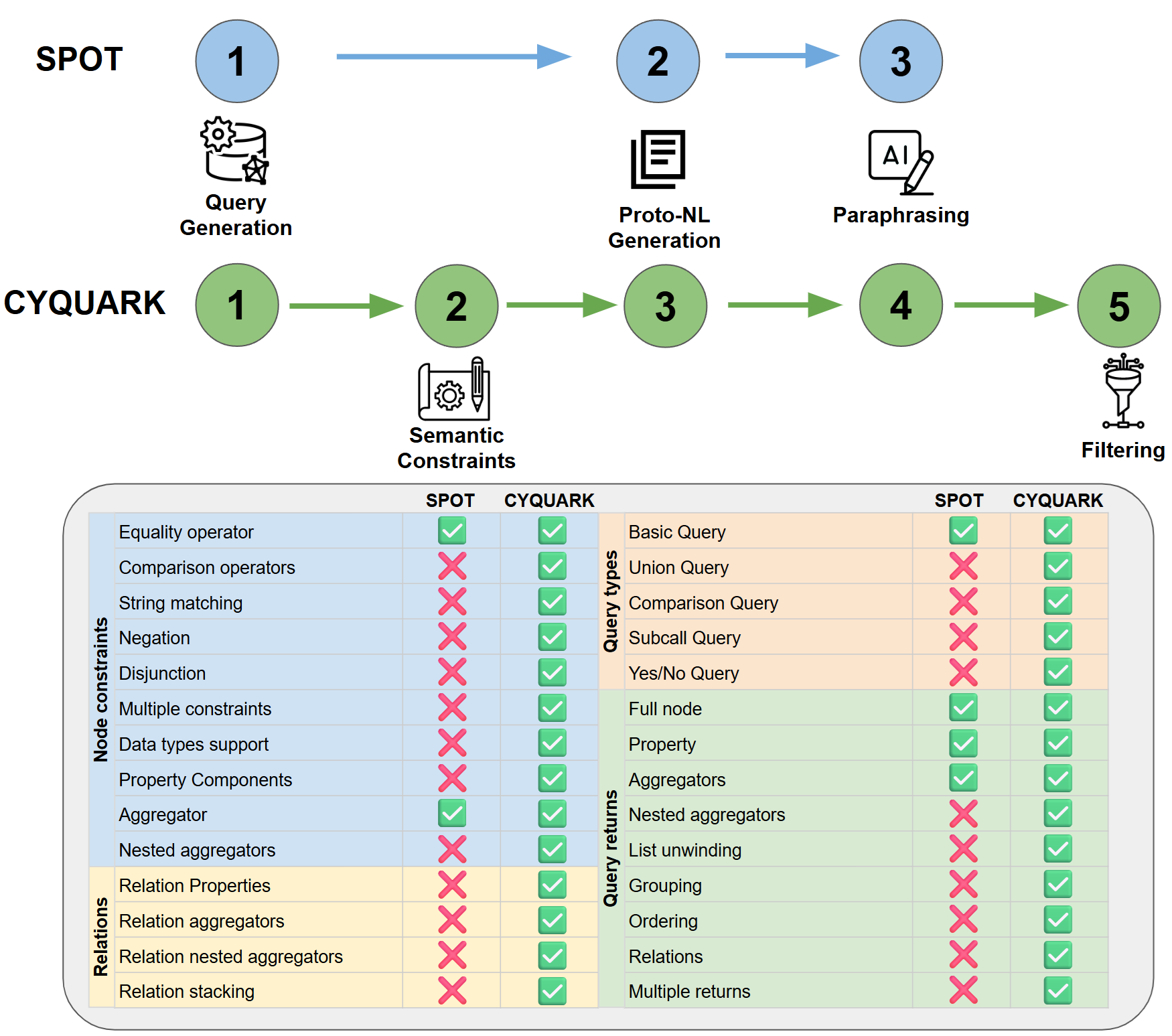}
  \caption{Comparison of the pipelines and supported constructs between SPOT and CYQUARK}
  \label{fig:cq_sp}
\end{figure*}

In this section we present our main contribution, CYQUARK, a method to automatically generate a diverse training set of annotated data pairs for a target PG. Our method, inspired by SPOT \citep{cazzaro-etal-2025-spot}, starts with a three stage pipeline approach, pattern generation and instantiation from target KG, finite-state proto-NL generation, and LLM paraphrasing, but adds two additional pipeline stages (Fig \ref{fig:cq_sp}, top).

The main difference between SPOT and CYQUARK is the expressivity of the generated queries. While SPOT generates only basic queries, CYQUARK covers most of the complex compositional constructs expressible with Cypher syntax. More precisely, in SPOT tree patterns have a rigid predefined structure which assumes that the query has a single answer node. All tree patterns generated by SPOT are rooted at this unique answer node and each path of the tree corresponds to a MATCH statement whose root is the expected query answer. 

The rigidity of the pattern structures produced by SPOT means that some important and common queries cannot be generated with the method. For example, consider the following query: ‘List the name of each teacher and the average number of students in their classes’ and its corresponding Cypher query: 
\begin{lstlisting}[style=cypher]
MATCH (x0:Teacher)-[:TEACHES]->(x1:Class)<-[:ATTENDS]-(x2:Student)
WITH x0, x1, COUNT(x2) AS c_x2
RETURN x0.name, AVG(c_x2)
\end{lstlisting}

This query cannot be generated by SPOT, since it exhibits a nested structure and multiple answer nodes, i.e. $x\_0 \text{ and }  c\_x2$. To address these complex constructs CYQUARK implements a more flexible pattern structure that no longer assumes a single answer node. Implementing these more general constructs involves not only changing the pattern generation step but also modifying the finite-state transducer compositional generation of the corresponding proto-NL realization. Furthermore, the added complexity requires adding additional verification steps to the overall generation pipeline architecture.

More precisely, enabling more complex patterns with multiple answer nodes and nested structures poses two important challenges that must be addressed in the generation process. 1) The added expressive power might lead to Cypher queries that have a correct logical meaning but that are semantically inconsistent.  2) The added query expressivity increases the intrinsic complexity and ambiguities of the corresponding natural language realizations (e.g. coreference ambiguities). 
To address these two challenges CYQUARK adds two additional stages to the generation pipeline (Fig \ref{fig:cq_sp}, top), a semantic consistency step (after Cypher query generation) and a filtering step (after LLM paraphrasing). Semantic constraints ensure semantic coherence of the generated Cypher query while the filtering step uses LLMs to ensure that the paraphrasing step did not introduce mistakes.

This section is organized as follows, section \ref{sec:complex_cypher_patt_gen} provides all the details of the patterns and constructs that can be generated with CYQUARK, while \ref{sec:semantic_constraints} and \ref{sec:logical_meaning_verification} provide more details about the semantic consistency and the filtering steps. 

\subsection{Complex Cypher Pattern Generation}
\label{sec:complex_cypher_patt_gen}

In contrast to SPOT, CYQUARK can generate a rich set of Cypher patterns covering most constructs of the Cypher syntax and representing a broad range of queries.
Figure \ref{fig:cq_sp} (bottom) presents a detailed comparison with SPOT. It is important to note that similarly to SPOT, CYQUARK's Cypher query and proto-NL generations are compositional. This means that the method can generate query/proto-NL pairs for every atomic pattern and every combination of these patterns, leading to an exponential growth in query expressivity.

We refer to a Basic Query type as a query that contains one or more \texttt{MATCH} clauses and a \texttt{RETURN}. Complex query types combine and extend basic queries. We can group complex queries into four classes (see Appendix \ref{app:query_types} for examples):

\textbullet \ \textbf{Union Query}. An Union Query combines the results of two or more Basic Queries. It applies a disjunctive logic, returning results that respect at least one of the constituent conditions.

\textbullet \ \textbf{Comparison Query}. A Comparison Query involves two Basic Queries and returns the result of one of them based on a comparison. For example, it may retrieve two persons and return the youngest one by comparing their ages.

\textbullet \ \textbf{Subcall Query}. A Subcall Query consists of a Basic Query that contains a nested Basic Query. This structure enables requests that cannot be expressed using a single Basic Query.

\textbullet \ \textbf{Yes/No Query}. A Yes/No Query involves a Basic Query modified to return True or False, evaluating an existence or condition statement.

CYQUARK can generate queries that use most of Cypher’s expressivity. For example, for node property constraints, CYQUARK supports equality and comparison operators (\texttt{>}, \texttt{<}, \texttt{>=}, \texttt{<=}), string-matching constraints, negation and regular expressions combining these operators over multiple properties. When generating feature constraints the generation process takes into account the native Cypher data-types (e.g. date, time, datetime, integer, float, duration, point coordinates, boolean, and lists). Appendix \ref{app:cyquark_features} describes all functionalities.

The generated node constraints can also result from applying functions to atomic feature values. These functions are called aggregators and include sum, count, average, min and max. For example, a query might request nodes for which a given feature value is above a given threshold. The generation also supports nested aggregator constraints. 

SPOT could only generate single returns, in contrast, CYQUARK is able to generate a wide range of return patterns, such as: returning a set of nodes, a subset of node properties, multiple returns of different types or the output of aggregator functions and nested aggregators. Furthermore, there is also flexibility in how returns can be ordered and grouped. All return types, constraints, aggregators, and nested aggregators are also supported for relationships. Besides, CYQUARK supports relationship stacking, which represents an OR disjunction between relationships on a single edge.

\subsection{Semantic Constraints}
\label{sec:semantic_constraints}

CYQUARK is capable of generating a vast array of rich and diverse queries and the generation process ensures that all of them are syntactically valid and that they can be executed against the target PG. However, syntactic correctness does not imply semantic coherence. The additional expressivity enabled by CYQUARK generation has a cost, it might introduce more noisy (i.e. semantically incoherent samples). To address this problem we add an additional component to the generation pipeline that filters queries based on a set of semantic constraint rules. These semantic constraints can be directly instantiated from the target graph schema. For example, an average aggregator function should only be applied to numeric data-types.

\subsection{Filtering}
\label{sec:logical_meaning_verification}

The finite-state process that generates the proto-NL for each generated Cypher query was carefully designed to be as unambiguous as possible. However, the expressivity (e.g. complex return patterns and nested structures) enabled by CYQUARK means that some language ambiguities are unavoidable. As a result the LLM paraphrasing step becomes harder and it might introduce some errors. More precisely, it might generate a more fluent natural language sentence (as compared to the proto-NL) but it might fail to preserve the logical meaning of the query. To address this, we add an additional step to the generation pipeline which uses an LLM as a filter. More precisely, we prompt an LLM to determine whether the natural language sentence generated in the paraphrasing stage preserves or not the meaning expressed in the Cypher query. 

\section{Experiments}

We test CYQUARK on four Text-To-Cypher benchmarks spanning $19$ graphs. We operate in a zero-shot scenario, meaning that we run CYQUARK with default settings and without any human intervention, thus no modifications are made to cater to the specifics of the different Knowledge Graphs. This ensures a fair comparison with out-of-the-box LLMs, which also operate in a zero-shot manner. However, note that specializing the settings and prompts based on the target Knowledge Graph is surely beneficial for the final parser. Indeed, the library is designed with maximum flexibility and customizability, allowing it to generate data specifically adapted to a particular Knowledge Graph.

\subsection{Benchmarks}
We test CYQUARK on the following benchmarks:

\textbullet\ \textsc{\textbf{MindTheQuery}} \citep{chauhan-etal-2025-mind}: A recently released benchmark, we run evaluations on all the graphs in the test set: Bloom (BL), Entity Resolution (ER), Contact Tracing (CT), Healthcare Analytics (HC), Women World Cup 2019 (WW).

\textbullet\ \textsc{\textbf{CypherBench}} \citep{feng-etal-2025-cypherbench}: This work focuses on building property graphs from RDF graphs and in particular from Wikidata. We consider all the graphs assigned to the test set: Company (CO), Fictional Character (FC), Flight Accident (FA), Geography (GE), Movie (MO), Nba (NB), Politics (PO).

\textbullet\ \textsc{\textbf{Text2GQL-Bench}} \citep{lyu2026text2gqlbenchtextgraphquery}: We consider all the graphs assigned to the test set: Financial Payment (FP), Information Technology (IT), Geography Graph (GG), Manufacturing Bill of Materials (MB), Manufacturing Production Process (MP), Social Network Twitter (SN).

\textbullet\ \textsc{\textbf{Zograscope}} \citep{cazzaro-etal-2025-zograscope}: A benchmark focused on a crime investigation graph, it comes with three test partitions: iid (IID), compositional (COMP), length (LEN).

\subsection{Models}

\begin{table*}[h]
\begin{center}
\begin{tabular}{lccccc|c}
\toprule
\textbf{Model} & \textbf{BL} & \textbf{ER} & \textbf{CT} & \textbf{HC} & \textbf{WW} & \textbf{AVG.}  \\ \midrule

Qwen3.5-4B & $84.07$ & $59.68$ & $45.88$ & $49.25$ & $59.41$ & $59.66$  \\
Qwen3.5-9B & $80.08$ & $60.83$ & $49.07$ & $61.21$ & $51.00$ & $60.44$  \\
Qwen3.6-27B  & $93.67$ & $66.62$ & $62.92$ & $82.78$ & $75.12$ & $76.22$  \\

\arrayrulecolor{gray}\midrule

GPT-5.5 & $98.17$ & $77.97$ & $77.84$ & $95.06$ & $89.39$ & $87.67$  \\
Claude Opus 4.7 & $97.72$ & $78.87$ & $77.05$ & $92.86$ & $91.01$ & $87.50$  \\

\arrayrulecolor{gray}\midrule

SPOT-4B & $51.00$ & $29.35$ & $35.25$ & $25.07$ & $40.90$ & $36.31$  \\

\arrayrulecolor{gray}\midrule

CYQUARK-0.8B & $89.30$ & $80.40$ & $63.55$ & $82.96$ & $77.44$ & $78.73$  \\
CYQUARK-4B & $91.26$ & $83.97$ & $66.75$ & $85.13$ & $80.29$ & $81.48$  \\

\arrayrulecolor{black}
\bottomrule
\end{tabular}
\end{center}
\caption{Execution accuracy results on MindTheQuery}
\label{tab:main_results_mindthequery}
\end{table*}

\begin{table*}[h]
\begin{center}
\begin{tabular}{lcccccc|c}
\toprule
\textbf{Model} & \textbf{FP} & \textbf{IT} & \textbf{GG} & \textbf{MB} & \textbf{MP} & \textbf{SN} & \textbf{AVG.}  \\ \midrule

Qwen3.5-4B & $24.12$ & $26.93$ & $39.75$ & $35.80$ & $44.14$ & $31.84$ & $33.76$ \\
Qwen3.5-9B & $29.63$ & $30.13$ & $43.66$ & $29.92$ & $45.18$ & $34.73$ & $35.54$  \\
Qwen3.6-27B  & $40.25$ & $46.61$ & $52.15$ & $46.91$ & $57.96$ & $55.27$ & $49.86$  \\

\arrayrulecolor{gray}\midrule

GPT-5.5 & $53.63$ & $67.39$ & $68.43$ & $66.14$ & $74.98$ & $61.36$ & $65.32$  \\
Claude Opus 4.7 & $48.43$ & $71.09$ & $61.90$ & $66.18$ & $69.18$ & $63.21$ & $63.33$  \\

\arrayrulecolor{gray}\midrule

SPOT-4B & $12.05$ & $16.51$ & $17.27$ & $15.46$ & $17.96$ & $13.74$ & $15.50$  \\

\arrayrulecolor{gray}\midrule

CYQUARK-0.8B & $59.95$ & $57.62$ & $46.89$ & $48.82$ & $60.27$ & $41.84$ & $54.49$  \\
CYQUARK-4B & $63.06$ & $62.11$ & $52.55$ & $53.46$ & $61.87$ & $44.31$ & $56.23$  \\

\arrayrulecolor{black}
\bottomrule
\end{tabular}
\end{center}
\caption{Execution accuracy results on Text2GQL-Bench}
\label{tab:main_results_mindthequery}
\end{table*}

We utilize the data generated by CYQUARK to fine-tune two small models of the Qwen family \citep{qwen35blog}, specifically \textit{Qwen3.5-0.8B} and \textit{Qwen3.5-4B}. For both paraphrasing and filtering we employ \textit{Qwen3.5-9B} using the prompts reported in Appendix \ref{app:paraph_prompt} and \ref{app:filter_prompt}. We report the number of generated samples and their overlap in Appendix \ref{app:gen_samples_stats}, along with examples of CYQUARK generations in Appendix \ref{sec:gen_examples}.

We compare CYQUARK with SPOT (with the same LLM models for fine-tuning and paraphrasing) and the de-facto alternative method when no training data is available for a parser: i.e. prompting LLMs in a zero shot fashion. We run the models without reasoning enabled. We construct a prompt (reported in Appendix \ref{app:texttocypher_prompt}) that provides the knowledge graph schema along with instructions to generate the Cypher query. We employ open-weights models from the Qwen family to have a direct comparison, specifically \textit{Qwen3.5-4B}, \textit{Qwen3.5-9B} and \textit{Qwen3.6-27B}. 
Additionally, we also compare with two proprietary closed source models: \textit{GPT-5.5} and Claude Opus 4.7. 
Although their exact parameter counts are undisclosed, they are generally estimated to be two orders of magnitude larger than the models we fine-tune using CYQUARK’s data. Besides, in Section \ref{sec:llm_data_comparison} we compare with models fine-tuned with alternative data generation pipelines based on LLMs.

\subsection{Macro-F1 Execution Accuracy}
\label{app:macro_f1}

\begin{table*}[h]
\begin{center}
\begin{tabular}{lccccccc|c}
\toprule
\textbf{Model} & \textbf{CO} & \textbf{FC} & \textbf{FA} & \textbf{GE} & \textbf{MO} & \textbf{NB} & \textbf{PO} & \textbf{AVG.}  \\ \midrule

Qwen3.5-4B & $19.48$ & $25.94$ & $33.91$ & $33.49$ & $33.65$ & $41.72$ & $24.51$ & $30.39$ \\
Qwen3.5-9B & $26.20$ & $35.90$ & $57.45$ & $41.12$ & $29.77$ & $42.28$ & $25.11$ & $36.83$  \\
Qwen3.6-27B  & $54.30$ & $65.96$ & $75.35$ & $64.68$ & $61.50$ & $76.74$ & $53.28$ & $64.54$ \\

\arrayrulecolor{gray}\midrule

GPT-5.5 & $89.32$ & $88.29$ & $92.17$ & $79.34$ & $88.71$ & $96.32$ & $87.37$ & $88.79$ \\
Claude Opus 4.7 & $87.57$ & $91.79$ & $91.72$ & $80.02$ & $87.97$ & $91.68$ & $89.32$ & $88.58$  \\

\arrayrulecolor{gray}\midrule

SPOT-4B & $49.59$ & $44.79$ & $47.58$ & $14.27$ & $46.18$ & $43.53$ & $45.21$ & $41.59$  \\

\arrayrulecolor{gray}\midrule

CYQUARK-0.8B & $81.06$ & $76.81$ & $86.37$ & $76.18$ & $84.94$ & $82.40$ & $80.12$ & $81.13$ \\
CYQUARK-4B & $81.79$ & $77.77$ & $87.96$ & $79.79$ & $86.92$ & $84.03$ & $82.70$ & $83.00$ \\

\arrayrulecolor{black}
\bottomrule
\end{tabular}
\end{center}
\caption{Execution accuracy results on CypherBench}
\label{tab:main_results_cypherbench}
\end{table*}

\begin{table*}[h]
\begin{center}
\begin{tabular}{lccc|c}
\toprule
\textbf{Model} & \textbf{IID} & \textbf{COMP} & \textbf{LEN} & \textbf{AVG.}  \\ \midrule

Qwen3.5-4B & $33.37$ & $31.50$ & $18.31$ & $27.73$  \\
Qwen3.5-9B & $41.99$ & $33.26$ & $16.80$ & $30.68$  \\
Qwen3.6-27B  & $50.06$ & $53.80$ & $38.21$ & $47.26$  \\

\arrayrulecolor{gray}\midrule

GPT-5.5 & $67.28$ & $72.01$ & $57.76$ & $65.68$   \\
Claude Opus 4.7 & $73.86$ & $75.09$ & $63.75$ & $70.90$   \\

\arrayrulecolor{gray}\midrule

SPOT-4B & $78.68$ & $77.36$ & $66.83$ & $74.29$  \\
    
\arrayrulecolor{gray}\midrule

CYQUARK-0.8B & $75.79$ & $80.90$ & $68.18$ & $74.96$   \\
CYQUARK-4B & $81.58$ & $81.09$ & $70.82$ & $77.83$   \\

\arrayrulecolor{black}
\bottomrule
\end{tabular}
\end{center}
\caption{Execution accuracy results on Zograscope}
\label{tab:main_results_mindthequery}
\end{table*}

Exact execution accuracy compares the result sets returned by the gold and predicted queries, assigning a correct score only when they match exactly. We instead use macro-F1, which gives partial matches a non-zero score. This is because exact execution accuracy might be excessively harsh, particularly because samples in the test sets, especially those generated by LLMs, may have results that do not exactly reflect the natural language question. Consider the following example from the MP test.

Question: \textit{Identify equipment resources that are required by manufacturing operations which are part of work orders marked as 'Completed'}

Query:\begin{lstlisting}[style=cypher]
MATCH (wo:WorkOrder)-[:TRIGGERS]->(mo:ManufacturingOperation)-[r:REQUIRES]->(er:EquipmentResource) 
WHERE wo.status = 'Completed' 
RETURN DISTINCT er.EquipmentResource_id, er.type
\end{lstlisting}

The question asks to identify equipment resources, and the query does so by returning the IDs of the equipment resources. However, the query also returns the equipment resource types. This information is not requested in the question and, while not necessarily incorrect, is somewhat arbitrary. If a model predicts a correct query that returns only the equipment resource IDs, it would be considered incorrect under exact execution accuracy. With macro-F1, however, it would receive at least a partially correct score. We have chosen to use macro-F1 rather than micro-F1 to give equal importance to each query. This is because queries can return significantly different numbers of results, affecting micro-F1.

Furthermore, for some properties, we consider both returning the entire node or the property to be correct. In the example above, the question asks for equipment resources, returning either the entire equipment resource node or just its ID would be acceptable. Our evaluation treats both cases equivalently. Note that this only applies to a few specific properties that essentially identify a node, such as IDs and names. If the question were instead asking for the type of equipment resource, only returning the property er.type would be considered correct, whereas returning the entire node would be counted as incorrect. We release our evaluation script alongside the code. From each test set we remove queries that return empty results or encounter execution errors. Appendix \ref{app:exp_metrics} reports standard deviation, confidence intervals and execution percentage.

\section{Results and discussion}

\begin{table*}[]
\begin{center}
\begin{tabular}{lcccc}
\toprule
\textbf{Model} & \textbf{MTQ} & \textbf{T2G} & \textbf{CBE} & \textbf{ZOG}\\ \midrule

CYQUARK-4B & $81.48$ & $56.23$ & $83.00$ & $77.83$  \\
- w.o. filtering & $79.23$ & $49.70$ & $80.34$ & $73.76$  \\
- w.o. semantic constraints & $80.53$ & $55.07$ & $80.96$ & $74.81$  \\
- w.o. both & $77.59$ & $48.25$ & $78.86$ & $70.76$   \\

\arrayrulecolor{black}
\bottomrule
\end{tabular}
\end{center}
\caption{Results ablating filtering, semantic constraints and both. The average for each benchmark is reported.}
\label{tab:filtering}
\end{table*}

\begin{table*}[h]
\begin{center}
\begin{tabular}{lccccccc}
\toprule
& \multicolumn{2}{c}{\textbf{Mind The Query}} & \multicolumn{2}{c}{\textbf{Text2GQL-Bench}} & \multicolumn{2}{c}{\textbf{CypherBench}} & \multicolumn{1}{c}{\textbf{Hetionet}} \\
\textbf{Model} & \textbf{ER} & \textbf{HC} & \textbf{FP} & \textbf{MP} & \textbf{MO} & \textbf{NB} & \\ \midrule

MTQ-4B & $83.50$ & $77.92$ & $34.09$ & $28.77$ & $24.66$ & $29.35$ & $-$  \\
SyntheT2C-4B & $-$ & $-$ & $-$ & $-$ & $-$ & $-$ & $37.89$  \\
CYQUARK-4B & $83.97$ & $85.13$ & $63.06$ & $61.87$ & $86.92$ & $84.03$ & $51.70$  \\

\arrayrulecolor{black}
\bottomrule
\end{tabular}
\end{center}
\caption{Execution accuracy comparison with LLM data generation pipelines}
\label{tab:llm_datagens}
\end{table*}

Tables 1, 2, 3 and 4 show results across the four Text-To-Cypher benchmarks. The first observation is that, as expected, the performance of the different models in a zero-shot scenario is highly dependent on their size. In particular, the performance of small models improves as we increase the size going from $37.94$ for the 4B model to $41.21$ for the 9B model and $60.67$ for the 27B (average performance across all benchmarks). While performance improves, even a 27B model exhibits weak zero-shot performance. 

The large models work very well for the CypherBench and MindTheQuery benchmarks but have weaker performance on Text2GQL-Bench and Zograscope. When looking at the performance of Claude versus GPT-5.5, there does not seem to be a significant difference. Claude is better for some benchmarks and GPT-5.5 for others, but on average over the four benchmarks Claude has $78.58$ accuracy and GPT-5.5 $78.52$, less than a $0.1$ difference.

If we focus on the results that we obtain by fine-tuning small LLMs with our synthetic data generation approach we observe a significant increase in performance. A 4B LLM with CYQUARK data doubles the average accuracy rate (i.e. $74.25$ versus $37.94$). Even a tiny 0.8B LLM fine-tuned with CYQUARK data has a $71.51$ average accuracy. With CYQUARK the performance of the small LLMs becomes very close to that of the most powerful proprietary LLMs, $74.25$ CYQUARK versus $78.58$ and $78.52$ of Claude and GPT-5 respectively. 
If we focus on the comparison with SPOT, CYQUARK is clearly superior, achieving substantial improvements across the board. Essentially, using CYQUARK it is possible to obtain a free, small locally deployable LLM with performance that is comparable to that of expensive models that charge for each single query. The cost of the computational resources required to generate the data and fine-tune the model is negligible and would quickly get amortized as more queries are processed.

\subsection{Error analysis}

We conducted an error analysis of the predictions made by CYQUARK. Queries with only one node were the easiest, with only $20$\% of them containing errors, whereas queries with multiple nodes were generally more complex and difficult. $40$\% of queries with five nodes contained errors. Another potential source of complexity is the use of aggregators, which were present in $47$\% of the queries that contained errors. However, this relationship may also be due to the fact that queries featuring aggregators are generally more complex. Furthermore, we examined the number of objects to be returned. When a query required only one return, the prediction contained an additional returned object in only $2$\% of cases. A higher number of returns was associated with a higher likelihood of either omitting or adding a return. Queries with four required returns had a $30$\% return mismatch rate.
    
To get a better understanding of the paraphrasing and filtering steps of our data generation pipeline, we conducted a manual evaluation study with human annotators experts on the Cypher language. We report the details of the study in Appendix \ref{app:paraph_quality}. Overall, we found that the paraphrasing step perfectly preserved the meaning of the Cypher query in $65$\% of cases. The filtering step achieved $80$\% recall and $66$\% precision in detecting paraphrasing errors. The high recall means that it can successfully remove most mistakes.

\subsection{Ablations}

Table \ref{tab:filtering} reports the average results for each benchmark for CYQUARK models with the filtering, the semantic constraints and both ablated (full breakdown in Appendix \ref{app:filtering_breakdown}).

Both steps consistently improve performance, yielding a $7.98$\% increase in the case of Text2GQL-Bench. The increase in performance for individual datasets varies, as high as $10$ percentage points for some datasets. Out of the two, the filtering seems to be the one with the biggest impact.

\subsection{LLM Data Generation}
\label{sec:llm_data_comparison}

We compare CYQUARK with two LLM based pipelines for generating training data: \textsc{MindTheQuery} (MTQ) and \textsc{SyntheT2C} \citep{zhong-etal-2025-synthet2c}. For the former, which was used to build the corresponding benchmark, we run the code released by the authors employing GPT-5.5 selecting six graphs from three of the benchmarks. The latter approach was developed on Hetionet \citep{10.7554/eLife.26726}, a biomedical KG, and its prompts contain graph specific instructions that make the approach difficult to transfer to other KGs. We utilize the generated data released by the authors.

The results are reported in Table \ref{tab:llm_datagens} (in Appendix \ref{app:equal_size_llms} we report results where CYQUARK is trained with the same exact number of samples). CYQUARK achieves considerably better performance overall. Only on \textsc{MindTheQuery} the results are comparable, which is expected since those benchmarks were generated using the same LLM pipeline. On the remaining graphs, however, MTQ fails to generalize effectively. LLM based data generation pipelines rely on the prompts to steer the generation of Cypher query templates, making it difficult to produce a diverse set of queries. A key strength of CYQUARK is its ability to generate a wide variety of query structures, which enables it to perform well across a broad range of KGs.

\section{Conclusion}

Property Graphs have emerged as powerful data models but accessing them via conversational interfaces requires robust Text-To-Cypher parsers. While LLMs can power these parsers, data sovereignty requirements often necessitate smaller, locally deployable models. Since small LLMs struggle with zero-shot performance, fine-tuning them with annotated data is necessary, yet acquiring such data remains a significant bottleneck.

Prior work addressed this by proposing a method to automatically generate training data for a target PG. However, this method was lacking in Cypher syntax coverage. This paper addresses this critical shortcoming by expanding the range of expressivity of the generated queries while making sure that the enriched expressivity does not compromise the quality of the generated data. We conducted extensive experiments on all the main Text-To-Cypher benchmarks. Our results show that our method can bring the performance of small locally deployable LLMs very close to that of proprietary LLMs. 

\section*{Acknowledgments}

This project has received funding from the European Research Council (ERC) under the European Union's Horizon 2020 research and innovation programme under grant agreement No 853459. The authors gratefully acknowledge the computer resources at ARTEMISA, funded by the European Union ERDF and Comunitat Valenciana as well as the technical support provided by the Instituto de Física Corpuscular, IFIC (CSIC-UV). This research is supported by a recognition 2021SGR-Cat (01266 LQMC) from AGAUR (Generalitat de Catalunya).

\section*{Limitations}

One limitation of CYQUARK is that it only supports English for the natural language questions and Cypher for the query language. This being said, because of its modular design, extending the tool to work with other graph query languages should be relatively easy. Generalizing to another graph language will involve adjusting the tree pattern generation process to reflect the syntax of the target language. Analogously, generalizing to other input languages would involve modifying the finite-state process that generates proto-NL from decorated tree-patterns, this second type of generalization would probably be more challenging but still feasible, also depending on how closely related the language is to English.

Furthermore as stated in the paper the process of paraphrasing as well as filtering could be improved by more complex approaches.

Our approach successfully generates complex tree patterns and translates them into both Cypher and natural language. However, real world user interactions with PGs are seldom limited to a single query, they are usually conversational and iterative. We believe that future research should focus on expanding this method to generate training data for multi round semantic parsing. This would shift the focus from one off queries to a continuous dialogue based interaction, better reflecting how users actually communicate their needs.

\section*{Risks and Ethical Considerations}

Our proposed method aims to facilitate more efficient information accessing to graph data. While we do not identify immediate ethical risks or safety concerns inherent in this work, we acknowledge that any system utilizing automatically generated data requires careful implementation. To prevent the propagation of factual inaccuracies or existing biases within Knowledge Graphs, we recommend that such systems be deployed with human oversight.


\bibliography{custom,anthology-1,anthology-2}

\appendix

\clearpage
\section{Experimental Details}

We fine-tune our models on a single A100 GPU, and all reported results are averaged over five runs with different random seeds. The seeds used are $1231$, $1232$, $1233$, $1234$ and $1235$. We use the PyTorch training framework (version $2.10.0$) with Python $3.11.14$ and the Transformers library (version $5.5.4$). We did minimal hyperparameter tuning in preliminary experiments, and all reported runs use the same configuration: a learning rate of $0.000005$, a weight decay of $0.01$, a single training epoch, and an effective batch size of $16$. The effective batch size is achieved using a per-device batch size of $2$ with gradient accumulation of $8$ steps for \textit{Qwen3.5-4B}, and a per-device batch size of $8$ with gradient accumulation of $2$ steps for \textit{Qwen3.5-0.8B}. We use the Adafactor optimizer and perform full precision fine-tuning. We release the training script in the associated repo.

\section{Paraphrasing Quality}
\label{app:paraph_quality}

To get a better understanding of the paraphrasing and filtering steps of our data generation pipeline, we conducted a manual evaluation study with human annotators experts on the Cypher language. We employed 2 annotators who had a 96\% agreement rate, disagreements were resolved by discussions. We generated $100$ samples with CYQUARK across all graphs, covering all different query types. The experiment was as follows: we showed the annotator the Cypher query and the corresponding natural language realization produced by our pipeline just before the filtering stage (i.e. after LLM paraphrasing). Then we asked the annotator to decide if the natural language query preserved the logical meaning of the Cypher query or not, essentially the annotator played the role of a logical meaning preservation verifier. Overall $65$\% of the generated natural language queries were correct.

Then we run the samples through our filtering stage and used the gold expert generated annotations to evaluate the precision and recall of the LLM filter when predicting incorrect paraphrases. The filter had an $80$ \% recall and a $66$ \% precision in predicting paraphrasing errors. The high recall means that it can successfully remove most mistakes. Precision is lower (i.e. it might wrongly remove good samples) but this is not a problem since we can always generate more samples.  

\clearpage  

\section{Ablations Breakdown}
\label{app:filtering_breakdown}

\begin{table}[ht]
\centering
\begin{tabular}{lccccccc}
\toprule
& \multicolumn{7}{c}{\textbf{Mind The Query}} \\
\textbf{Model} & \textbf{BL} & \textbf{ER} & \textbf{CT} & \textbf{HC} & \textbf{WW} & &  \\
\midrule

CYQUARK-0.8B & $89.30$ & $80.40$ & $63.55$ & $82.96$ & $77.44$  \\
- w.o. filtering & $90.27$ & $74.73$ & $61.25$ & $81.14$ & $77.82$   \\
\arrayrulecolor{gray}\midrule
CYQUARK-4B & $91.26$ & $83.97$ & $66.75$ & $85.13$ & $80.29$   \\
- w.o. filtering & $87.85$ & $80.14$ & $66.35$ & $84.04$ & $77.76$  \\
- w.o. semantic constraints & $90.92$ & $83.43$ & $65.40$ & $82.81$ & $80.08$  \\
- w.o. both & $87.23$ & $79.28$ & $64.40$ & $80.89$ & $76.17$   \\

\arrayrulecolor{black}

\midrule
& \multicolumn{7}{c}{\textbf{Text2GQL-Bench}}\\
 & \textbf{FP} & \textbf{IT} & \textbf{GG} & \textbf{MB} & \textbf{MP} & \textbf{SN} &  \\
\midrule
CYQUARK-0.8B & $59.95$ & $57.62$ & $46.89$ & $48.82$ & $60.27$ & $41.84$   \\
- w.o. filtering & $52.51$ & $56.27$ & $40.36$ & $47.71$ & $56.92$ & $30.47$   \\
\arrayrulecolor{gray}\midrule
CYQUARK-4B & $63.06$ & $62.11$ & $52.55$ & $53.46$ & $61.87$ & $44.31$  \\
- w.o. filtering & $54.69$ & $57.63$ & $46.61$ & $48.95$ & $57.53$ & $32.79$  \\
- w.o. semantic constraints & $62.91$ & $59.69$ & $51.53$ & $52.97$ & $59.42$ & $43.87$ \\
- w.o. both & $52.59$ & $55.73$ & $44.19$ & $48.51$ & $56.86$ & $31.62$  \\
\arrayrulecolor{black}

\midrule
& \multicolumn{7}{c}{\textbf{CypherBench}}\\
 & \textbf{CO} & \textbf{FC} & \textbf{FA} & \textbf{GE} & \textbf{MO} & \textbf{NB} & \textbf{PO} \\
\midrule
CYQUARK-0.8B & $81.06$ & $76.81$ & $86.37$ & $76.18$ & $84.94$ & $82.40$ & $80.12$  \\
- w.o. filtering & $76.66$ & $74.35$ & $83.73$ & $74.16$ & $81.65$ & $78.11$ & $77.77$  \\
\arrayrulecolor{gray}\midrule
CYQUARK-4B & $81.79$ & $77.77$ & $87.96$ & $79.79$ & $86.92$ & $84.03$ & $82.70$  \\
- w.o. filtering & $76.97$ & $76.58$ & $86.92$ & $77.51$ & $85.06$ & $79.61$ & $79.70$   \\
- w.o. semantic constraints & $80.30$ & $75.65$ & $85.23$ & $76.32$ & $84.95$ & $82.4$ & $81.86$  \\
- w.o. both & $76.91$ & $73.96$ & $84.62$ & $74.87$ & $84.56$ & $79.91$ & $77.17$   \\
\arrayrulecolor{black}

\midrule
& \multicolumn{7}{c}{\textbf{ZOGRASCOPE}}\\
 & \textbf{IID} & \textbf{COMP} & \textbf{LEN} &  &  & &  \\
\midrule
CYQUARK-0.8B & $75.79$ & $80.90$ & $68.18$   \\
- w.o. filtering & $74.32$ & $76.93$ & $67.69$ \\
\arrayrulecolor{gray}\midrule
CYQUARK-4B & $81.58$ & $81.09$ & $70.82$    \\
- w.o. filtering & $77.94$ & $77.01$ & $66.34$    \\
- w.o. semantic constraints & $78.02$ & $78.25$ & $68.17$   \\
- w.o. both & $74.43$ & $74.32$ & $63.53$    \\
\arrayrulecolor{black}
\bottomrule

\end{tabular}
\caption{Breakdown of the results ablating filtering, semantic constraints and both.}
\label{tab:ablation_cyquark}
\end{table}

\clearpage
\section{Experiment metrics}
\label{app:exp_metrics}

\begin{table}[ht]
\centering
\begin{tabular}{lccccccc}
\toprule
& \multicolumn{7}{c}{\textbf{Mind The Query}} \\
\textbf{Model} & \textbf{BL} & \textbf{ER} & \textbf{CT} & \textbf{HC} & \textbf{WW} & &  \\
\midrule
SPOT-4B        & $\pm 0.45$ & $\pm 1.24$ & $\pm 0.58$ & $\pm 0.66$ & $\pm 1.07$ &  &  \\
\arrayrulecolor{gray}\midrule
CYQUARK-0.8B    & $\pm 1.72$ & $\pm 0.53$ & $\pm 0.12$ & $\pm 0.52$ & $\pm 0.79$ &  &  \\
CYQUARK-4B      & $\pm 1.09$ & $\pm 1.51$ & $\pm 0.96$ & $\pm 1.11$ & $\pm 1.23$ &  &  \\
\arrayrulecolor{black}

\midrule
& \multicolumn{7}{c}{\textbf{Text2GQL-Bench}}\\
 & \textbf{FP} & \textbf{IT} & \textbf{GG} & \textbf{MB} & \textbf{MP} & \textbf{SN} &  \\
\midrule
SPOT-4B        & $\pm 1.60$ & $\pm 0.64$ & $\pm 1.95$ & $\pm 1.51$ & $\pm 1.39$ & $\pm 1.03$ &  \\
\arrayrulecolor{gray}\midrule
CYQUARK-0.8B    & $\pm 1.05$ & $\pm 0.53$ & $\pm 0.35$ & $\pm 1.39$ & $\pm 0.91$ & $\pm 1.16$ &  \\
CYQUARK-4B      & $\pm 1.93$ & $\pm 1.40$ & $\pm 0.89$ & $\pm 0.63$ & $\pm 0.44$ & $\pm 0.96$ &  \\
\arrayrulecolor{black}

\midrule
& \multicolumn{7}{c}{\textbf{CypherBench}}\\
 & \textbf{CO} & \textbf{FC} & \textbf{FA} & \textbf{GE} & \textbf{MO} & \textbf{NB} & \textbf{PO} \\
\midrule
SPOT-4B        & $\pm 0.99$ & $\pm 0.80$ & $\pm 1.46$ & $\pm 0.90$ & $\pm 0.55$ & $\pm 1.40$ & $\pm 0.43$ \\
\arrayrulecolor{gray}\midrule
CYQUARK-0.8B    & $\pm 1.08$ & $\pm 0.35$ & $\pm 2.29$ & $\pm 1.61$ & $\pm 1.31$ & $\pm 1.89$ & $\pm 0.84$  \\
CYQUARK-4B      & $\pm 1.23$ & $\pm 1.81$ & $\pm 1.19$ & $\pm 1.34$ & $\pm 1.19$ & $\pm 1.05$ & $\pm 1.76$ \\
\arrayrulecolor{black}

\midrule
& \multicolumn{7}{c}{\textbf{ZOGRASCOPE}}\\
 & \textbf{IID} & \textbf{COMP} & \textbf{LEN} &  &  & &  \\
\midrule
SPOT-4B        & $\pm 1.13$ & $\pm 0.80$ & $\pm 0.73$ &  &  &  &  \\
\arrayrulecolor{gray}\midrule
CYQUARK-0.8B    & $\pm 0.45$ & $\pm 1.04$ & $\pm 0.53$ &  &  &  &  \\
CYQUARK-4B      & $\pm 1.28$ & $\pm 1.99$ & $\pm 1.71$ &  &  &  &  \\
\arrayrulecolor{black}
\bottomrule

\end{tabular}
\caption{Standard deviation on fine-tuned models.}
\label{tab:standard_dev_table}
\end{table}

\begin{table*}[h]
\centering
\begin{tabular}{lccccccc}
\toprule

& \multicolumn{5}{c}{\textbf{Mind The Query}} \\
\textbf{Model} & \textbf{BL} & \textbf{ER} & \textbf{CT} & \textbf{HC} & \textbf{WW} \\ \midrule

Qwen3.5-4B & $\pm 4.00$ & $\pm 3.27$ & $\pm 3.61$ & $\pm 3.51$ & $\pm 3.18$  \\
Qwen3.5-9B & $\pm 4.36$ & $\pm 3.26$ & $\pm 3.62$ & $\pm 3.42$ & $\pm 3.24$   \\
Qwen3.6-27B  & $\pm 2.66$ & $\pm 3.15$ & $\pm 3.50$ & $\pm 2.65$ & $\pm 2.80$   \\

\arrayrulecolor{gray}\midrule

GPT-5.5 & $\pm 1.46$ & $\pm 2.77$ & $\pm 3.01$ & $\pm 1.52$ & $\pm 1.99$   \\
Claude Opus 4.7 & $\pm 1.63$ & $\pm 2.73$ & $\pm 3.05$ & $\pm 1.81$ & $\pm 1.85$   \\

\arrayrulecolor{gray}\midrule

SPOT-4B & $\pm 5.46$ & $\pm 3.04$ & $\pm 3.46$ & $\pm 3.05$ & $\pm 3.18$   \\

\arrayrulecolor{gray}\midrule

CYQUARK-0.8B & $\pm 3.38$ & $\pm 2.65$ & $\pm 3.49$ & $\pm 2.64$ & $\pm 2.71$  \\
CYQUARK-4B & $\pm 3.09$ & $\pm 2.45$ & $\pm 3.42$ & $\pm 2.50$ & $\pm 2.58$   \\

\arrayrulecolor{black}
\midrule

& \multicolumn{5}{c}{\textbf{Text2GQL-Bench}} \\
\textbf{Model} & \textbf{FP} & \textbf{IT} & \textbf{GG} & \textbf{MB} & \textbf{MP} & \textbf{SN}   \\ \midrule

Qwen3.5-4B & $\pm 3.16$ & $\pm 3.31$ & $\pm 4.74$ & $\pm 4.33$ & $\pm 4.20$ & $\pm 3.89$ \\
Qwen3.5-9B & $\pm 3.38$ & $\pm 3.42$ & $\pm 4.81$ & $\pm 4.14$ & $\pm 4.21$ & $\pm 3.98$   \\
Qwen3.6-27B  & $\pm 3.63$ & $\pm 3.72$ & $\pm 4.84$ & $\pm 4.51$ & $\pm 4.18$ & $\pm 4.16$   \\

\arrayrulecolor{gray}\midrule

GPT-5.5 & $\pm 3.69$ & $\pm 3.50$ & $\pm 4.50$ & $\pm 4.27$ & $\pm 3.66$ & $\pm 4.07$  \\
Claude Opus 4.7 & $\pm 3.69$ & $\pm 3.38$ & $\pm 4.71$ & $\pm 4.27$ & $\pm 3.91$ & $\pm 4.03$  \\

\arrayrulecolor{gray}\midrule

SPOT-4B & $\pm 2.41$ & $\pm 2.77$ & $\pm 3.66$ & $\pm 3.26$ & $\pm 3.25$ & $\pm 2.88$ \\

\arrayrulecolor{gray}\midrule

CYQUARK-0.8B & $\pm 3.62$ & $\pm 3.68$ & $\pm 4.84$ & $\pm 4.51$ & $\pm 4.14$ & $\pm 4.12$  \\
CYQUARK-4B & $\pm 3.57$ & $\pm 3.62$ & $\pm 4.84$ & $\pm 4.50$ & $\pm 4.11$ & $\pm 4.15$  \\

\arrayrulecolor{black}
\midrule

\multicolumn{7}{c}{\textbf{CypherBench}} \\
\textbf{Model} & \textbf{CO} & \textbf{FC} & \textbf{FA} & \textbf{GE} & \textbf{MO} & \textbf{NB} & \textbf{PO}  \\ \midrule

Qwen3.5-4B & $\pm 4.17$ & $\pm 4.40$ & $\pm 6.75$ & $\pm 4.87$ & $\pm 4.62$ & $\pm 5.88$ & $\pm 4.27$ \\
Qwen3.5-9B & $\pm 4.63$ & $\pm 4.82$ & $\pm 7.05$ & $\pm 5.08$ & $\pm 4.48$ & $\pm 5.89$ & $\pm 4.30$  \\
Qwen3.6-27B  & $\pm 5.24$ & $\pm 4.76$ & $\pm 6.14$ & $\pm 4.93$ & $\pm 4.76$ & $\pm 5.04$ & $\pm 4.95$  \\

\arrayrulecolor{gray}\midrule

GPT-5.5 & $\pm 3.25$ & $\pm 3.23$ & $\pm 3.83$ & $\pm 4.18$ & $\pm 3.10$ & $\pm 2.25$ & $\pm 3.30$  \\
Claude Opus 4.7 & $\pm 3.47$ & $\pm 2.76$ & $\pm 3.93$ & $\pm 4.12$ & $\pm 3.18$ & $\pm 3.29$ & $\pm 3.07$ \\

\arrayrulecolor{gray}\midrule

SPOT-4B & $\pm 5.26$ & $\pm 4.99$ & $\pm 7.12$ & $\pm 3.61$ & $\pm 4.88$ & $\pm 5.91$ & $\pm 4.94$   \\

\arrayrulecolor{gray}\midrule

CYQUARK-0.8B & $\pm 4.12$ & $ \pm 4.24$ & $\pm 4.89$ & $\pm 4.39$ & $\pm 3.50$ & $\pm 4.54$ & $\pm 3.96$  \\
CYQUARK-4B & $\pm 4.06$ & $\pm 4.18$ & $\pm 4.64$ & $\pm 4.14$ & $\pm 3.30$ & $\pm 4.37$ & $\pm 3.75$  \\

\arrayrulecolor{black}
\midrule
& \multicolumn{3}{c}{\textbf{ZOGRASCOPE}} \\
\textbf{Model} & \textbf{IID} & \textbf{COMP} & \textbf{LEN} \\ \midrule

Qwen3.5-4B & $\pm 3.33$ & $\pm 2.48$ & $\pm 2.14$  \\
Qwen3.5-9B & $\pm 3.49$ & $\pm 2.51$ & $\pm 2.07$   \\
Qwen3.6-27B  & $\pm 3.54$ & $\pm 2.66$ & $\pm 2.69$   \\

\arrayrulecolor{gray}\midrule

GPT-5.5 & $\pm 3.32$ & $\pm 2.40$ & $\pm 2.73$   \\
Claude Opus 4.7 & $\pm 3.11$ & $\pm 2.31$ & $\pm 2.66$    \\

\arrayrulecolor{gray}\midrule

SPOT-4B & $\pm 2.90$ & $\pm 2.23$ & $\pm 2.61$   \\
    
\arrayrulecolor{gray}\midrule

CYQUARK-0.8B & $\pm 3.03$ & $\pm 2.10$ & $\pm 2.58$    \\
CYQUARK-4B & $\pm 2.74$ & $\pm 2.09$ & $\pm 2.52$    \\

\arrayrulecolor{black}
\bottomrule

\end{tabular}
\caption{$95$\% Wald confidence interval for all results.}
\label{tab:ci_prec_table}
\end{table*}

\begin{table*}[h]
\centering
\begin{tabular}{lccccccc}
\toprule
& \multicolumn{7}{c}{\textbf{Mind The Query}} \\
\textbf{Model} & \textbf{BL} & \textbf{ER} & \textbf{CT} & \textbf{HC} & \textbf{WW} & &  \\
\midrule
Qwen3.5-4B      & $94.72$ & $73.90$ & $73.46$ & $82.78$ & $83.19$ &  &  \\
Qwen3.5-9B      & $92.24$ & $81.90$ & $74.28$ & $82.78$ & $80.89$ &  &  \\
Qwen3.6-27B      & $98.45$ & $85.61$ & $72.78$ &  $97.56$ & $86.79$ &  &  \\
\arrayrulecolor{gray}\midrule
GPT-5.5      & $100$ & $99.65$ & $93.57$ & $100$ & $98.80$ &  &  \\
Claude Opus 4.7      & $100$ & $99.07$ & $86.87$ & $100$ & $99.67$ &  &  \\
\arrayrulecolor{gray}\midrule
SPOT-4B        & $98.76$ & $97.68$ & $86.46$ & $96.91$ & $94.43$ &  &  \\
\arrayrulecolor{gray}\midrule
CYQUARK-0.8B    & $99.69$ & $97.10$ & $94.94$ & $99.49$ & $91.27$ &  &  \\
CYQUARK-4B      & $99.93$ & $97.49$ & $93.98$ & $99.54$ & $94.08$ &  &  \\
\arrayrulecolor{black}

\midrule
& \multicolumn{7}{c}{\textbf{Text2GQL-Bench}}\\
 & \textbf{FP} & \textbf{IT} & \textbf{GG} & \textbf{MB} & \textbf{MP} & \textbf{SN} &  \\
\midrule
Qwen3.5-4B      & $61.88$ & $68.16$ & $71.64$ & $71.97$ & $75.23$ & $70.73$   &  \\
Qwen3.5-9B      & $67.99$ & $73.52$ & $82.64$ & $72.40$ & $77.84$ & $72.36$ &  \\
Qwen3.6-27B      & $92.60$ & $87.12$ & $87.04$ & $87.90$ & $89.94$ & $89.64$ &  \\
\arrayrulecolor{gray}\midrule
GPT-5.5      & $97.72$ & $91.03$ & $98.04$ & $98.94$ & $99.25$ & $97.09$ &  \\
Claude Opus 4.7      & $97.01$ & $98.99$ & $98.78$ & $99.15$ & $99.25$ & $96.54$ &  \\
\arrayrulecolor{gray}\midrule
SPOT-4B        & $88.62$ & $89.72$ & $92.66$ & $95.75$ & $92.92$ & $90.27$ &  \\
\arrayrulecolor{gray}\midrule
CYQUARK-0.8B    & $94.03$ & $95.22$ & $94.86$ & $96.39$ & $98.14$ & $97.45$ &  \\
CYQUARK-4B      & $96.93$ & $96.87$ & $95.49$ & $98.05$ & $98.57$ & $95.27$ &  \\
\arrayrulecolor{black}

\midrule
& \multicolumn{7}{c}{\textbf{CypherBench}}\\
 & \textbf{CO} & \textbf{FC} & \textbf{FA} & \textbf{GE} & \textbf{MO} & \textbf{NB} & \textbf{PO} \\
\midrule
Qwen3.5-4B      & $66.86$ & $77.95$ & $88.36$ & $73.13$ & $74.81$  &  $70.00$ &  $68.97$ \\
Qwen3.5-9B      & $70.32$ & $70.60$ & $87.30$ & $75.35$ & $66.58$ & $68.89$ & $65.13$ \\
Qwen3.6-27B      & $93.37$ & $91.60$ & $93.12$ & $87.26$ & $92.52$ & $94.81$ & $88.97$ \\
\arrayrulecolor{gray}\midrule
GPT-5.5      & $97.98$ & $99.74$ & $100$ & $98.61$ & $98.25$ & $99.63$ & $96.92$ \\
Claude Opus 4.7      & $98.85$ & $100$ & $100$ & $98.89$ & $99.00$ & $98.89$ &  $98.46$ \\
\arrayrulecolor{gray}\midrule
SPOT-4B        & $99.13$ & $99.21$ & $98.94$ & $95.84$ & $96.51$ & $96.67$ & $98.72$ \\
\arrayrulecolor{gray}\midrule
CYQUARK-0.8B    & $97.98$ & $100$ & $100$ & $100$ & $100$ & $99.26$ & $98.72$ \\
CYQUARK-4B      & $98.67$ & $99.89$ & $100$ & $99.72$ & $99.75$ & $99.63$ & $99.69$ \\
\arrayrulecolor{black}

\midrule
& \multicolumn{7}{c}{\textbf{ZOGRASCOPE}}\\
 & \textbf{IID} & \textbf{COMP} & \textbf{LEN} &  &  & &  \\
\midrule
Qwen3.5-4B      & $97.65$ & $92.81$ & $91.54$ & & &  &  \\
Qwen3.5-9B      & $97.39$ & $92.81$ & $91.30$ &  & &  &  \\
Qwen3.6-27B      & $97.26$ & $97.55$ & $95.69$ &  & &  &  \\
\arrayrulecolor{gray}\midrule
GPT-5.5      & $93.10$ & $94.66$ & $91.38$ &  & &  &  \\
Claude Opus 4.7 & $97.26$ & $97.48$ & $96.88$ &  & &  &  \\
\arrayrulecolor{gray}\midrule
SPOT-4B        & $99.76$ & $99.26$ & $98.43$ &  &  &  &  \\
\arrayrulecolor{gray}\midrule
CYQUARK-0.8B    & $100$ & $99.85$ & $99.92$ &  &  &  &  \\
CYQUARK-4B      & $99.95$ & $99.97$ & $99.95$ &  &  &  &  \\
\arrayrulecolor{black}
\bottomrule

\end{tabular}
\caption{Execution percentage on all benchmarks. Execution percentage is the percentage of queries that execute successfully on the KG without errors. If a query executes, it effectively rules out syntax errors in the Cypher structure as well as schema-linking mistakes. Consequently, if it contains some errors, these are purely semantic.}
\label{tab:exec_prec_table}
\end{table*}

\clearpage
\section{Generated Samples Statistics}
\label{app:gen_samples_stats}

\begin{table}[ht]
\centering
\begin{tabular}{lccccccc}
\toprule
& \multicolumn{5}{c}{\textbf{Mind The Query}} & \multicolumn{2}{c}{\textbf{ZOGRASCOPE}} \\
\textbf{Model} & \textbf{BL} & \textbf{ER} & \textbf{CT} & \textbf{HC} & \textbf{WW} & &   \\
\midrule
CYQUARK before filtering    & $300958$ & $80939$ & $212043$ & $137162$ & $189572$ &  & $691296$ \\
CYQUARK      & $170691$ & $55172$ & $144144$ & $87027$ & $130063$ &  & $512974$ \\
\arrayrulecolor{gray}\midrule
SPOT        & $350000$ & $258540$ & $254667$ & $350000$ & $350000$ &  & $350000$  \\
\arrayrulecolor{black}

\midrule
& \multicolumn{7}{c}{\textbf{Text2GQL-Bench}}\\
 & \textbf{FP} & \textbf{IT} & \textbf{GG} & \textbf{MB} & \textbf{MP} & \textbf{SN} &  \\
\midrule
CYQUARK before filtering   & $92485$ & $369404$ & $225562$ & $429819$ & $250887$ & $507209$ &  \\
CYQUARK      & $57127$ & $230430$ & $158807$ & $237723$ & $165072$ & $272550$ &  \\
\arrayrulecolor{gray}\midrule
SPOT        & $350000$ & $233265$ & $97411$ & $350000$ & $350000$ & $350000$ &  \\
\arrayrulecolor{black}

\midrule
& \multicolumn{7}{c}{\textbf{CypherBench}}\\
 & \textbf{CO} & \textbf{FC} & \textbf{FA} & \textbf{GE} & \textbf{MO} & \textbf{NB} & \textbf{PO} \\
\midrule
CYQUARK before filtering    & $162056$ & $248827$ & $190220$ & $164992$ & $51244$ & $119014$ & $178342$  \\
CYQUARK      & $102415$ & $177694$ & $121996$ & $100252$ & $35342$ & $75981$ & $109070$ \\
\arrayrulecolor{gray}\midrule
SPOT        & $350000$ & $350000$ & $350000$ & $350000$ & $350000$ & $350000$ & $350000$ \\

\arrayrulecolor{black}

\bottomrule

\end{tabular}
\caption{Number of generated samples for each graph.}
\label{tab:gene_samples_stats}
\end{table}

We computed the template-overlap between the Cypher queries generated by CYQUARK and those in the test sets of the benchmarks (note that, due to the different formats in which the queries are written, the results might not be entirely accurate). The overlap is $39.73$\% on MindTheQuery, $2.59$\% on Text2GQL-Bench, $25.89$\% on CypherBench and $22.63$\% on Zograscope.
Note that synthetic data generation methods aim to generate samples that are as realistic and diverse as possible and that could potentially be seen at test time (i.e. provided by a user in a real-world scenario, by a benchmark in a simulated setting). Therefore, this is a desirable characteristic of an effective synthetic data generation method. The fact that CYQUARK is grounded in the target knowledge graph and utilizes its actual values is a strength of the method. The goal is producing the best possible parser for the specific target knowledge graph.

\clearpage

\section{LLM Data Generation Results}
\label{app:equal_size_llms}

\begin{table}[h]
\begin{center}
\begin{tabular}{lccccccc}
\toprule
& \multicolumn{2}{c}{\textbf{Mind The Query}} & \multicolumn{2}{c}{\textbf{Text2GQL-Bench}} & \multicolumn{2}{c}{\textbf{CypherBench}} & \multicolumn{1}{c}{\textbf{Hetionet}} \\
\textbf{Model} & \textbf{ER} & \textbf{HC} & \textbf{FP} & \textbf{MP} & \textbf{MO} & \textbf{NB} & \\ \midrule

MTQ-4B & $83.50$ & $77.92$ & $34.09$ & $28.77$ & $24.66$ & $29.35$ & $-$  \\
SyntheT2C-4B & $-$ & $-$ & $-$ & $-$ & $-$ & $-$ & $37.89$  \\
CYQUARK-4B & $82.88$ & $83.20$ & $61.25$ & $58.13$ & $84.26$ & $82.29$ & $49.23$  \\

\arrayrulecolor{black}
\bottomrule
\end{tabular}
\end{center}
\caption{Execution accuracy comparison with LLM data generation pipelines when CYQUARK is trained with the exact same number of samples generated by the LLM methods for each graph.}
\label{tab:equalsize_llm_datagens}
\end{table}

\section{Graph Sizes}
\label{app:graph_sizes}

\begin{table}[ht]
\centering
\begin{tabular}{llcc}
\toprule
& \textbf{Graph} & \textbf{\# of nodes}  & \textbf{\# of relations} \\
\midrule
\footnotesize\textit{Mind The Query}    & Bloom (BL) & $30960$ & $29731$ \\
& Entity Resolution (ER) & $1237$ & $1819$  \\
& Contact Tracing (CT) & $5615$ & $15130$  \\
& Healthcare Analytics (HC) & $11381$ & $61453$  \\
& Women World Cup 2019 (WW) & $2486$ & $14799$  \\
\arrayrulecolor{gray}\midrule
\footnotesize\textit{Text2GQL-Bench} & Financial Payment (FP) & $8420$ & $17184$   \\
& Information Technology (IT) & $815$ & $3035$  \\
& Geography Graph (GG) & $1100$ & $5697$  \\
& Manufacturing Bill of Materials (MB) & $1500$ & $1770$  \\
& Manufacturing Production Process (MP) & $1030$ & $1595$  \\
& Social Network Twitter (SN) & $8450$ & $52608$  \\
\arrayrulecolor{gray}\midrule
\footnotesize\textit{CypherBench} & Company (CO)& $581306$ & $299581$  \\
& Fictional Character (FC) & $28915$ & $40097$  \\
& Flight Accident (FA) & $1683$ & $2212$  \\
& Geography (GE) & $773489$ & $903559$  \\
& Movie (MO) & $459393$ & $1892202$  \\
& Nba (NB) & $4327$ & $18991$  \\
& Politics (PO) & $885188$ & $1548416$  \\
\arrayrulecolor{gray}\midrule
\footnotesize\textit{ZOGRASCOPE} & Pole & $61521$ & $105840$  \\
\arrayrulecolor{black}

\bottomrule

\end{tabular}
\caption{Graph nodes and relations statistics}
\label{tab:graph_sizes}
\end{table}

\clearpage

\section{CYQUARK Features}  
\label{app:cyquark_features}

\lstset{
  basicstyle=\ttfamily\footnotesize,
  breaklines=true,
  breakatwhitespace=true,
  columns=fullflexible,
  keepspaces=true,
  showstringspaces=false
}

\begin{table}[!h]
\centering
\begin{tabularx}{\textwidth}{l l X p{3cm}}
\toprule
& \textbf{Feature} & \textbf{Query example} & \textbf{Options} \\
\midrule

\multicolumn{1}{l}{\footnotesize\textit{Nodes}} 
& Equality 
& \lstinline|(x0:Person WHERE x0.nationality = 'German')| 
& - \\

& Comparison operators
& \lstinline|(x0:Person WHERE x0.birth_date > date('1980-11-24'))| 
& \texttt{>}, \texttt{<}, \texttt{>=}, \texttt{<=} \\

& String Matching 
& \lstinline|(x0:Book WHERE x0.title CONTAINS 'Dinosaurs')| 
& \texttt{STARTS WITH},  \texttt{ENDS WITH}, \texttt{CONTAINS} \\

& Negation 
& \lstinline|(x0:Person WHERE NOT x0.nationality = 'German')| 
& - \\

& Multiple constraints
& \lstinline|(x0:Person WHERE x0.nationality = 'German' AND x0.birth_date.year > 1980)| 
& - \\

& Disjunction
& \lstinline|(x0:Person WHERE x0.nationality = 'German' or x0.nationality = 'Spanish')| 
& - \\

& Data types support
& \lstinline|(x0:Place WHERE x0.coordinates = Point(\{x: 51.2312, y:4.4185\}))| 
& integer, float, time, dates, duration, point, boolean \\

& Components 
& \lstinline|(x0:Person WHERE x0.birth_date.year > 1980)| 
& year, month, day, ... \\

\multicolumn{1}{l}{\footnotesize\textit{}} 
& Aggregators 
& \lstinline|(x0:Company)
RETURN x0
ORDER BY x0.launch_year ASC
LIMIT 1
| 
& \texttt{COUNT}, \texttt{MAX}, \texttt{MIN}, \texttt{TOP\_N}, \texttt{AVG}, \texttt{SUM} \\

& Nested aggregators
& \lstinline|WITH COUNT(DISTINCT x0) AS z0_0
WHERE AVG(z0_0) > 2 | 
& any in aggregators \\

\arrayrulecolor{gray}
\midrule

\multicolumn{1}{l}{\footnotesize\textit{Relations}}  
& Properties
& \lstinline|(x0)-[r:Founded WHERE r.year = 2010]-(x1)| 
& any node property constraints \\

& Aggregators
& \lstinline|WHERE COUNT(r) > 5 | 
& \texttt{COUNT}, \texttt{MAX}, \texttt{MIN}, \texttt{TOP\_N}, \texttt{AVG}, \texttt{SUM} \\

& Nested Aggregators
& \lstinline|WITH COUNT(r) AS z0_0
WHERE AVG(z0_0) > 3| 
& any in aggregators \\

& Stacking
& \lstinline|(x0)-[:FRIEND_OF\|COLLEAGUE_OF]-(x1)| 
& - \\

\midrule

\multicolumn{1}{l}{\footnotesize\textit{Returns}} 
& Node 
& \lstinline|RETURN x0| 
& - \\

& Property
& \lstinline|RETURN x0.age| 
& - \\

& Aggregators
& \lstinline|RETURN COUNT(x0)| 
& \texttt{COUNT}, \texttt{MAX}, \texttt{MIN}, \texttt{AVG}, \texttt{SUM} \\

& Nested aggregators
& \lstinline|WITH COUNT(x0) AS z0 RETURN AVG(z0)| 
& any in aggregators \\

& List unwinding
& \lstinline|UNWIND x0.nicknames AS uw RETURN uw | 
& - \\

& Ordering
& \lstinline|RETURN x0 ORDER BY x0.date ASC| 
& - \\

& Relations
& \lstinline|MATCH (x0)-[r]-(x1) RETURN r| 
& - \\

& Multiple returns
& \lstinline|RETURN x0.age, x1, x2.nationality| 
& - \\

\arrayrulecolor{black}
\bottomrule
\end{tabularx}
\caption{Supported CYQUARK Cypher features}
\label{tab:query_features}
\end{table}

\clearpage

\section{Query Types}
\label{app:query_types}

\begin{table}[!h]
\begin{center}
\begin{tabular}{@{}ll@{}}

\toprule

Type &\parbox[t]{1.8\columnwidth}{
\begin{small}
BASIC QUERY
\end{small}
}\\
NL &\parbox[t]{1.8\columnwidth}{
\begin{small}
 \textit{List the name of each teacher and the average number of students in their classes}
\end{small}
}\\
Query & 
\begin{minipage}{0.8\linewidth}
\begin{lstlisting}[style=cypher]
MATCH (x0:Teacher)-[:TEACHES]->(x1:Class)<-[:ATTENDS]-(x2:Student)
WITH x0, x1, COUNT(x2) AS c_x2
RETURN x0.name, AVG(c_x2)
\end{lstlisting} \end{minipage} \\

\midrule

Type &\parbox[t]{1.8\columnwidth}{
\begin{small}
UNION QUERY
\end{small}
}\\
NL &\parbox[t]{1.8\columnwidth}{
\begin{small}
 \textit{What are the capitals of the countries that are located in Asia or have an area of at least 1,787,000 square kilometers?}
\end{small}
}\\
Query & 
\begin{minipage}{0.8\linewidth}
\begin{lstlisting}[style=cypher]
CALL () {
  MATCH (x0:Country)-[:locatedIn]-(x1:Continent WHERE x1.name = "Asia")
  RETURN x0
  UNION
  MATCH (x0:Country WHERE x0.area_km2 >= 1787000)
  RETURN x0
}
WITH DISTINCT x0
RETURN x0.capital

\end{lstlisting} \end{minipage} \\

\midrule
Type &\parbox[t]{1.8\columnwidth}{
\begin{small}
COMPARISON QUERY
\end{small}
}\\
NL &\parbox[t]{1.8\columnwidth}{
\begin{small}
 \textit{Who is younger, the director named John or the consultant named Brad?}
\end{small}
}\\
Query & 
\begin{minipage}{0.8\linewidth}
\begin{lstlisting}[style=cypher]
MATCH (x0:Director WHERE x0.name = "John")
MATCH (y0:Consultant WHERE y0.name = "Brad")
RETURN CASE
  WHEN x0.age < y0.age THEN x0
  ELSE y0
END AS answer

\end{lstlisting} \end{minipage} \\

\midrule
Type &\parbox[t]{1.8\columnwidth}{
\begin{small}
SUBCALL QUERY
\end{small}
}\\
NL &\parbox[t]{1.8\columnwidth}{
\begin{small}
 \textit{Return each person together with the youngest person they have met}
\end{small}
}\\
Query & 
\begin{minipage}{0.8\linewidth}
\begin{lstlisting}[style=cypher]
MATCH (x0:Person)
CALL (x0) {
MATCH (x0:Person)-[:met]-(x1:Person)
RETURN x1 as sb1
ORDER BY x1.age
LIMIT 1
}
RETURN x0, sb1
\end{lstlisting} \end{minipage} \\

\midrule
Type &\parbox[t]{1.8\columnwidth}{
\begin{small}
YES/NO QUERY
\end{small}
}\\
NL &\parbox[t]{1.8\columnwidth}{
\begin{small}
 \textit{Is there a country whose capital is Windhoek?}
\end{small}
}\\
Query & 
\begin{minipage}{0.8\linewidth}
\begin{lstlisting}[style=cypher]
RETURN EXISTS {
MATCH (x0:Country_ge WHERE x0.capital = "Windhoek")
}

\end{lstlisting} \end{minipage} \\

\bottomrule
\end{tabular}
\end{center}
\caption{Examples of different types of Cypher queries.}
\label{tab:query_Types}
\end{table}

\clearpage

\section{Generation Examples}
\label{sec:gen_examples}
\begin{table}[!h]
\begin{center}
\begin{tabular}{@{}ll@{}}

\toprule

Proto-NL &\parbox[t]{1.8\columnwidth}{
\begin{small}
list name of each Company with launch year that is  1962 that 

-  operate in radio communications
\end{small}
}\\
NL &\parbox[t]{1.8\columnwidth}{
\begin{small}
 \textit{Identify the names of all companies established in 1962 that are currently active in the radio communications sector.}
\end{small}
}\\
Query & 
\begin{minipage}{0.8\linewidth}
\begin{lstlisting}[style=cypher]
MATCH (x0:Company WHERE x0.launch_year = 1962)-[:operatesIn]-(x1:Industry WHERE x1.name = "radio communications")
RETURN x0.name
\end{lstlisting} \end{minipage} \\

\midrule

Proto-NL &\parbox[t]{1.8\columnwidth}{
\begin{small}
what is the sum of number of deaths of  Flight Accident that 

-  involve Aircraft Model with height in meters that is  7.1 (list also the maximum of the Aircraft Model wingspan) that were produced by Aircraft Manufacturer with country that is  United States of America
\end{small}
}\\
NL &\parbox[t]{1.8\columnwidth}{
\begin{small}
 \textit{If we look at flight accidents involving aircraft from US manufacturers with a height of 7.1 meters, what is the sum of fatalities recorded, and what is the largest wingspan found in this specific group of models?}
\end{small}
}\\
Query & 
\begin{minipage}{0.8\linewidth}
\begin{lstlisting}[style=cypher]
MATCH (x0:FlightAccident)-[:involves]-(x1:AircraftModel WHERE x1.height_metre = 7.1)-[:manufacturedBy]-(x2:AircraftManufacturer WHERE x2.country = "United States of America")
WITH SUM(x0.number_of_deaths) as su_x0, MAX(x1.wingspan_metre) as ma_x1
RETURN su_x0, ma_x1
\end{lstlisting} \end{minipage} \\

\midrule

Proto-NL &\parbox[t]{1.8\columnwidth}{
\begin{small}
list year of founding date of each Country that are either

 Country with official language that is  Haitian Creole 
 
OR

 Country with founding date that is less or equal than 1949-05-09 and founding date that is greater or equal than 1939-03-14
\end{small}
}\\
NL &\parbox[t]{1.8\columnwidth}{
\begin{small}
 \textit{For countries founded anytime from March 14, 1939, up to May 9, 1949, or those that designate Haitian Creole as their official tongue, please provide their founding year.}
\end{small}
}\\
Query & 
\begin{minipage}{0.8\linewidth}
\begin{lstlisting}[style=cypher]
CALL () {
  MATCH (x0:Country WHERE "Haitian Creole" IN x0.official_language)
  RETURN x0
  UNION
  MATCH (x0:Country WHERE x0.founding_date <= date('1949-05-09') AND x0.founding_date >= date('1939-03-14'))
  RETURN x0
} WITH DISTINCT x0 RETURN x0.founding_date.year
\end{lstlisting} \end{minipage} \\

\midrule
Proto-NL &\parbox[t]{1.8\columnwidth}{
\begin{small}
Does exist  Asset with id that is  A00435 that 

-  is located at Location with name that is not  Lowehaven Office?
\end{small}
}\\
NL &\parbox[t]{1.8\columnwidth}{
\begin{small}
 \textit{Does the asset bearing the identifier A00435 have a location that differs from the Lowehaven Office?}
\end{small}
}\\
Query & 
\begin{minipage}{0.8\linewidth}
\begin{lstlisting}[style=cypher]
RETURN EXISTS {
MATCH (x0:ASSET WHERE x0.asset_id = "A00435")-[:LOCATED_AT]-(x1:LOCATION WHERE NOT x1.name = "Lowehaven Office") }
\end{lstlisting} \end{minipage} \\

\bottomrule
\end{tabular}
\end{center}
\caption{Examples of CYQUARK generations.}
\label{tab:cyquark_examples}
\end{table}

\onecolumn
\section{Paraphrasing Prompt}
\label{app:paraph_prompt}

\begin{lstlisting}[style=prompt, linewidth=0.99\linewidth]
### Role
You are an expert NLP specialist and query paraphrasing assistant. Your goal is to convert artificial, syntax-heavy sentences (derived from Cypher queries) into natural, fluent, and grammatically correct English questions or requests.
### Critical Logic Extraction
Before rewriting, mentally extract the **exact** list of entities, labels, relationship types, and filters from the input sentence.
- **WARNING**: Do not summarize. Do not generalize. Do not omit any specific values (dates, names, IDs, numbers). The output must logically entail exactly the same results as the input query.
### Input Peculiarities & Handling
The input sentences often contain specific artifacts from graph query generation. You must handle these as follows:
1. **Numerical/Alphabetic References**: If you see markers like `(1)`, `(X)` or similar references used to point to specific nodes, do not include them in the final output.. Treat them as internal pointers to understand the sentence.
2. **"And also the target" Phrasing**: If the sentence includes a clause like "and also the target," understand this as an instruction to apply new filters or specifications **to the very first entity mentioned** (the root node).
3. **Lists and Multiple Lines**: If the input appears as a bulleted list or split across multiple lines, convert this into a single, cohesive, flowing sentence.
4. **Minimum and maximum**: Usually used for ordering. Minimum of a date would be the earliest, first or oldest. Depending on object paraphrase with qualifiers such as shortest or biggest.
### Constraints & Rules
1. **Strict Semantic Integrity**:
   - The output must answer the **exact** same question as the input.
   - **Forbidden**: Adding external assumptions, omitting critical constraints, or merging distinct entities into vague groups.
   - **Mandatory**: Every specific entity, filter, and relationship mentioned in the original sentence must be represented explicitly or clearly in the paraphrase.
- Note that years can range even beyond 2026
2. **Natural Fluency**:
- The output must sound like a native human wrote it. Eliminate robotic phrasing, awkward "It is...", "The one that...", and forced grammar structures common in code-to-text generation.
3. **Maximum Variation**:
   - Use diverse synonyms for key terms (e.g., instead of "has a relationship with," use "is connected to," "relates to," "is linked with").
   - Change the sentence structure (e.g., transform active voice to passive, or reorder clauses).
   - Break down long logical chains into smoother, more conversational phrasing without losing steps.
   - It is absolutely important that the three sentences are as different as possible from one another. Particularly in the structure and order the things are mentioned. DO use logical leaps! 1) 2) and 3) must be different! Still list everything that must be in output exactly
   - All caps AND is a logical divider that introduces a new condition or relation independent from the one before, there can be multiple independent AND
   - either plus all caps OR is a disjunction between statements before and after the OR (the possible AND are part of the disjuncted statements)
4. **Output Format**:
   - Provide exactly **three** distinct paraphrased sentences.
   - Number them clearly as 1), 2), and 3).
   - **Crucial**: Ensure all three options preserve every detail from the original.
   - Immediately after the third option (3), add a single line containing only the tag: `[END]`.
   - Do not include explanations, introductions, JSON, or markdown code blocks around the text.
### Examples
**Example Sentence**:
How many Pages that 
- belong to Recipe (for each list also grouped by Recipe type) that require Ingredient (for each list also the average of how many of these the Ingredient per Recipe)
**Example Paraphrase**:
1) How many pages are dedicated to recipes, grouped by recipe type, and on average what is the number of ingredients involved for each recipe
2) For each recipe type, what is their total page length and mean number of ingredients for a single one
3) List the average number of ingredients used in a single recipe and the total count of pages they occupy, tallied per recipe type
**Example Sentence**:
what is the average duration of each Standup Comedy with duration that is greater or equal than 12 that 
- were sponsored by any Media Company that sponsored Standup Comedy with duration that is 18 (for each list also the Standup Comedy original language)
**Example Paraphrase**:
1) What is the average duration of stand-up comedy performances lasting at least 12 minutes that are sponsored by media companies that also sponsored standups of 18 minutes, list also the original language of those 18 minute shows
2) list the language spoken at 18 minute standups sponsored by a media company. For this media company also list the average length of sponsored standups longer than 12 minutes
3) If we look at one specific media company sponsor's 18-minute standup shows, what languages are spoken? And separately, what is the average length of all their other sponsored standups that go on for more than 12 minutes?
**Example Sentence**:
list name of each Province that 
- contain any Amusement Park (for each list also the minimum of the Amusement Park cost)
- is hosting - with end date that is null - Event with type that is sport
**Example Paraphrase**:
1) Among provinces hosting a sport event that has not finished yet, list their name along with the cost of the least expensive amusement park in it
2) What is the cost of the cheapest amusement park in each province that has an ongoing sport event? Also what is the name of each Province?
3) For every province where a sport event is still happening, list the province name and the cost of its cheapest amusement park
**Example Sentence**:
list age of each person named John Smith that 
- has written Book that has released Version with the sum of earning that is greater or equal than 10234
**Example Paraphrase**:
1) What is John Smith's age, given that he has authored a book earning more than 10,234?
2) For books which combined versions earnings is more than 10234, list the age of its author if it is named John Smith
3) Find the ages of every individual named John Smith that has earned at least 10234 from writing a book
### Task
Please paraphrase the following sentence following the rules above.
**Original Sentence**:
[[QUERY]]
\end{lstlisting}

\section{Filtering Prompt}
\label{app:filter_prompt}

\begin{lstlisting}[style=prompt, linewidth=0.99\linewidth]
Does the following query make sense?
Consider only the logical meaning. The syntax is correct, it is just a newer cypher version.
You have to consider if this is something someone would ask or it would just be confusing.
We also provide a schematic meaning of the query in natural language
answer with yes or no in between <sense> tags. You can include a very short explanation in between <explanation> tags.
Query:
[[QUESTION]]

[[QUERY]]
\end{lstlisting}

\section{Text-To-Cypher Prompt}
\label{app:texttocypher_prompt}

\begin{lstlisting}[style=prompt, linewidth=0.99\linewidth]
### Role
You are an expert Neo4j and Cypher specialist. Your task is to translate natural language questions into accurate, executable Cypher queries.

### Schema
You must strictly follow this graph schema. Only use the node labels, relationship types, and properties explicitly defined here:
[[SCHEMA]]

### Instructions
- Generate a valid Cypher query that can run directly in Neo4j without modification.
- Use only the labels, relationships, and properties defined in the schema.
- If a property requires operations (e.g., AVG, SUM, comparisons) and needs casting from the schema, apply the appropriate type casting function (e.g., toInteger(), toFloat()) before using it.
- Return exactly and only the data requested.
- Do not include explanations or comments.

### Output Format
Wrap the final Cypher query strictly between the following tags:
<query>
your Cypher query here
</query>

### Task
Convert the following natural language request into a Cypher query:
<question>
[[QUERY]]
</question>
\end{lstlisting}

\twocolumn

\end{document}